\documentclass[lettersize,journal]{IEEEtran}
\usepackage{amsmath,amsfonts}
\usepackage{algorithmic}
\usepackage{algorithm}
\usepackage{array}
\usepackage[caption=false,font=normalsize,labelfont=sf,textfont=sf]{subfig}
\usepackage{textcomp}
\usepackage{stfloats}
\usepackage{url}
\usepackage{verbatim}
\usepackage{graphicx}
\usepackage{cite}
\usepackage{enumerate}
\usepackage{amsmath}
\usepackage{color}
\begin{document}

\title{Single Image Super-Resolution Based on Global-Local Information Synergy}

\author{Nianzu~Qiao,~Lamei~Di,~and~Changyin~Sun}

\markboth{Journal of \LaTeX\ Class Files,~Vol.~14, No.~8, August~2022}%
{Shell \MakeLowercase{\textit{et al.}}: A Sample Article Using IEEEtran.cls for IEEE Journals}

\IEEEpubid{0000--0000/00\$00.00~\copyright~2022 IEEE}

\maketitle

\begin{abstract}
Although several image super-resolution solutions exist, they still face many challenges. CNN-based algorithms, despite the reduction in computational complexity, still need to improve their accuracy. While Transformer-based algorithms have higher accuracy, their ultra-high computational complexity makes them difficult to be accepted in practical applications. To overcome the existing challenges, a novel super-resolution reconstruction algorithm is proposed in this paper. The algorithm achieves a significant increase in accuracy through a unique design while maintaining a low complexity. The core of the algorithm lies in its cleverly designed Global-Local Information Extraction Module and Basic Block Module. By combining global and local information, the Global-Local Information Extraction Module aims to understand the image content more comprehensively so as to recover the global structure and local details in the image more accurately, which provides rich information support for the subsequent reconstruction process. Experimental results show that the comprehensive performance of the algorithm proposed in this paper is optimal, providing an efficient and practical new solution in the field of super-resolution reconstruction. 
\end{abstract}


\section{Introduction}
\IEEEPARstart{S}{uper-resolution} was proposed by Gerchberg et al. \cite{ref115} to improve the spatial resolution of optical systems. Nowadays, super-resolution reconstruction is widely used in medical imaging, intelligent displays, remote sensing imaging and other fields.

The early designs of super-resolution reconstruction algorithms relied heavily on traditional machine learning algorithms, including interpolation-based \cite{ref116,ref117}, reconstruction-based \cite{ref118,ref119}, and shallow learning-based \cite{ref120,ref121} super-resolution reconstruction algorithms. The rapid rise of deep learning has led researchers to nowadays focus on deep learning-based super-resolution reconstruction algorithms, which in turn has driven unprecedented performance results for this class of algorithms.

Deep learning algorithms in this field are categorized into three groups: image super-resolution reconstruction algorithms based on CNN, GAN and Transformer.

GAN-based image super-resolution reconstruction algorithm. Ledig et al. \cite{ref156} first applied GAN technique to atmospheric image super-resolution reconstruction and the algorithm was named Super-Resolution Generative Adversarial Network (SRGAN). Wang et al. \cite{ref157} unearthed an optimization algorithm for SRGAN. It is Enhanced Super-Resolution Generative Adversarial Network (ESRGAN). The algorithm designed by Sajjadi et al. \cite{ref158} adds a discriminator in the feature domain to enhance its performance. Park et al. \cite{ref159} developed a Super-Resolution with Feature Discrimination (SRFeat) model. Pan \cite{ref160} et al. used an asymptotic approach to construct layers of generators and discriminators. Wei et al. \cite{ref161} utilized the wavelet transform to input high-frequency components into the GAN network. The gradient loss method proposed by Ma et al. \cite{ref162} is able to bootstrap the model using structural information, thus improving its performance. Liang et al. \cite{ref163} explicitly differentiated between visual artifacts and realistic details and regularized the GAN training framework.
\IEEEpubidadjcol

Transformer-based image super-resolution reconstruction algorithm. Liang et al. \cite{ref165} uncovered the super-resolution reconstruction algorithm SwinIR based on Swin Transformer \cite{ref166}. Zhang et al. \cite{ref167} developed an Efficient Long-range Attention Network (ELAN). Based on SwinIR, Zhou et al. \cite{ref168} designed an optimized SRFormer algorithm.

CNN-based image super-resolution reconstruction algorithm. Dong et al. \cite{ref128,ref129} initialized a CNN-based super-resolution reconstruction technique (Super-Resolution Convolutional Neural Network, SRCNN). Based on the work of Simonyan et al. \cite{ref130}, Kim et al. \cite{ref131} unearthed the 20-layer Very Deep Super-Resolution Reconstruction (VDSR) algorithm. Mao et al. \cite{ref132} relied on the U-Net structure and designed the Residual Encoder- Decoder Network (RED-Net). Kim et al. \cite{ref133} and Tai et al. \cite{ref134,ref135} both employed recursive structures to achieve the goal of super-resolution reconstruction of images.Dong et al. \cite{ref136} attempted for the first time to use up-sampling at the end of the model to generate super-resolution images, which was an innovation that greatly improved the computational efficiency of the model. Shi et al. \cite{ref137} used the up-sampling method at the end of the model to generate super-resolution images, which was an innovation that greatly improved the computational efficiency of the model. \cite{ref137} explored a sub-pixel convolution module to improve the performance of super-resolution reconstruction, and Lim et al. \cite{ref138} developed a super-resolution reconstruction algorithm based on a multiscale improved residual structure. Ahn et al. \cite{ref139} proposed a model that fuses both local and global information. Tong et al. \cite{ref140} and Zhang et al. \cite{ref141} developed two different models based on the model structure of DenseNet \cite{ref142}, and each of them developed two different model structures based on DenseNet \cite{ref142}. based on the DenseNet \cite{ref142} model structure, and each developed two different super-resolution reconstruction algorithms. Zhang et al. \cite{ref143}, Mei et al. \cite{ref144,ref145}, and Magid et al. \cite{ref146} each combined the attention mechanism with a deep learning model to uncover a series of innovative super-resolution reconstruction algorithms. Wang et al. \cite{ref147} designed a sparse coding network model (Sparse Coding Network Model) based on the concept of sparse coding \cite{ref148}. Wang et al. \cite{ref147} designed the Sparse Coding based Network (SCN) based on the concept of sparse coding \cite{ref148}. Lai et al. \cite{ref149,ref150} uncovered an innovative super-resolution network architecture, which is the Laplacian Pyramid Super-Resolution Network (LapSRN). Haris et al. \cite{ref151} designed a Deep Back-Projection Network (DBPN). Haris et al. \cite{ref152} developed an improved RBPN network based on DBPN. Li et al. \cite{ref153} utilized a loop structure to create SRFBN. Kong et al. \cite{ref154} designed an efficient super-resolution reconstruction algorithm named Residual Local Feature Network (RLFN). Sun et al. \cite{ref155} developed a spatially-adaptive feature modulation network (SAFMN), which was designed to provide a spatially-adaptive feature modulation network. SAFMN) designed to perform super-resolution reconstruction efficiently.

GAN-based image super-resolution reconstruction algorithms focus on global perception, but do not perform well in detail reconstruction and are prone to artifacts and noise. This is an inherent problem of this type of algorithm, which is difficult to avoid. Transformer-based image super-resolution reconstruction algorithms are still in the preliminary development stage, facing challenges such as high computational complexity and complex model structure. CNN-based image super-resolution reconstruction algorithms have numerous advantages and currently perform optimally in terms of comprehensive performance (complexity and accuracy). However, there is still room for improvement in this class of algorithms. For example, CNN-based algorithms have certain deficiencies in long-range dependency capturing, which is a factor that contributes to the limited performance of this class of algorithms.

\section{The recommended approach}
In this paper, we will introduce an efficient deep learning algorithm for image super-resolution reconstruction based on global-local information and describe its core components. As shown in Fig. 1, the model consists of the following parts: the basic Block module and the global-local information extraction module. Specifically, in this paper, an $3\times 3$ convolutional layer is first responsible for converting the input low-resolution (LR) image into a feature space, which in turn generates shallow features. Subsequently, multiple stacked Block modules are utilized to generate finer deep features from these shallow features for super-resolution (SR) image reconstruction. These deep features are then passed through the Global-Local Information Extraction module to further extract their global and local information. Eventually, the PixelShuffle technique is applied to scale up the size of the feature map to generate the final output.

\begin{figure*}[t]
	\centering
	\includegraphics[width=6in]{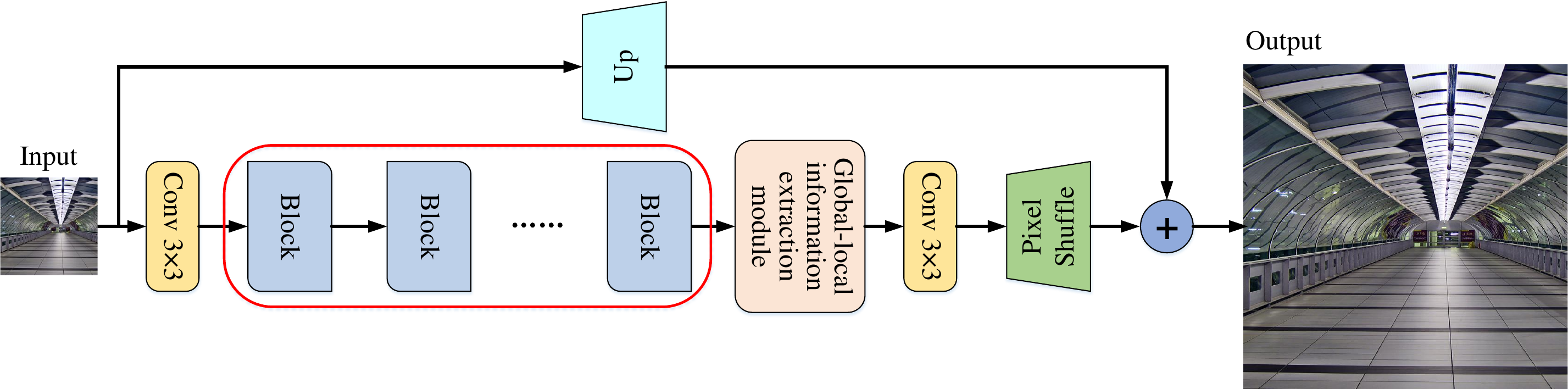}
	\caption{Schematic structure of deep learning algorithm for image super-resolution reconstruction based on global-local information.}
	\label{fig1}
\end{figure*}

In the next sections, this paper will delve into the technical details of the basic Block module and the Global-Local Information Extraction module. These modules play a crucial role in the super-resolution reconstruction algorithm of this paper, so understanding how they work is essential to better understand how the whole algorithm works.

\subsection{Basic Block Module}
The details of the basic Block module are shown in Fig. 2. Special attention should be paid to the fact that the basic Block module employs a method of feature map dot product instead of the traditional nonlinear activation function. This method accomplishes the fusion of channel information while realizing the activation function, thus providing a stronger nonlinear expression capability. Because of this innovative strategy, the basic Block module shows excellent performance in the overall algorithm and provides strong support for the final super-resolution image reconstruction.

\begin{figure*}[h]
	\centering
	\includegraphics[width=6.5in]{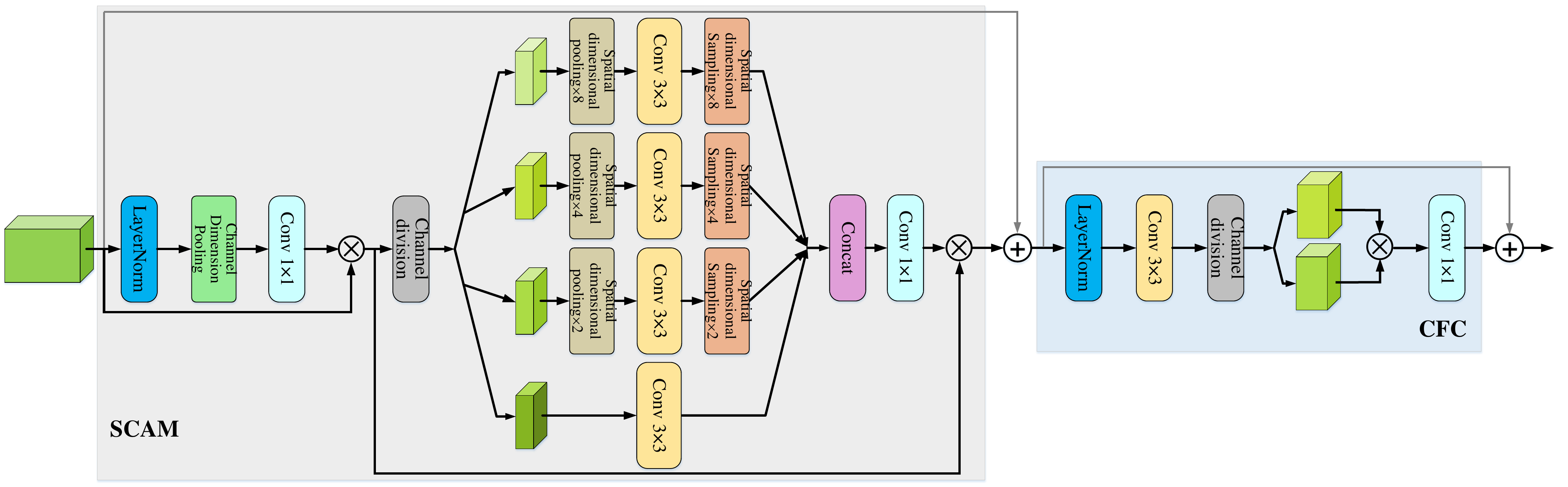}
	\caption{Diagram of Basic Block Module.}
	\label{fig2}
\end{figure*}

The basic Block module consists of two parts: the Spatial Channel Adaptive Modulation (SCAM) module and the Channel Fusion Convolution (CFC) module.

\subsubsection{SCAM Module}

Compared with the module proposed in the literature \cite{ref206}, this paper designs a new optimized module. This module not only considers the adaptive feature modulation in spatial dimension, but also introduces the adaptive feature modulation mechanism in channel dimension. As a result, the module is not only able to adaptively adjust the features in different spatial scales, but also able to modulate intelligently according to the feature differences between channels.

Specifically, the adaptive feature modulation mechanism for multi-scale spatial dimensions allows the module to capture and analyze image features at different spatial scales. As a result, the module can be more flexible in responding to details and structures at different scales in the image, thus extracting richer and more useful information.

And the adaptive feature modulation mechanism of the channel dimension enables the module to more accurately identify and extract the channel features that are crucial for HR image reconstruction, thus further improving the quality and efficiency of image reconstruction.

By combining the above two mechanisms, the optimized module designed in this paper aims to more deeply mine and utilize the features that are crucial for HR image reconstruction. As shown in Fig. 2, the specific process of the SCAM module includes channel adaptive modulation in the first part and spatial adaptive modulation in the second part. The specific process of the first part is:

\begin{itemize}
	\item First, the input feature map $X$ undergoes a normalization operation to generate the feature map $\hat{X}$.
	
	\item The feature map $\hat{X}$ is then transformed into a feature vector ${{\hat{X}}_{0}}$ by Avgpool of the channel dimensions.
	
	\item Subsequently, the feature vector ${{\hat{X}}_{0}}$ is subjected to $1\times 1$ convolution operation and the generated result is multiplied with the input feature map $X$ to obtain the feature map ${{\hat{X}}_{1}}$.
\end{itemize}

The mathematical expression of the first part is given in the following equation.

\begin{equation}
	{{\hat{X}}_{1}}=X*Conv1\times 1\left( P\left( L\left( X \right) \right) \right).
\end{equation}

Where $L$ stands for LayerNorm normalization operation. $P$ stands for the average pooling of channel dimensions.

The specific process for the second part is:

\begin{itemize}
	\item First, the feature map ${{\hat{X}}_{1}}$ is sliced into 4 sub-feature maps, and the spatial size of each sub-feature map is kept the same as the original feature map ${{\hat{X}}_{1}}$. The number of channels is reduced to $1/4$ of the original number.
	
	\item Then, 1 sub-feature map is randomly retained without any processing. And the remaining 3 sub-feature maps are implemented with $\times 2$, $\times 4$ and $\times 8$ times rate downsampling operations respectively.
	
	\item Immediately thereafter, four sub-feature maps of different sizes are processed separately by depth-separable convolution. Although these feature maps are processed by the same size convolution kernel, the perceptual field of each pixel varies depending on the feature map scale, with the smaller scale feature maps having larger perceptual fields. For subsequent stitching operations in the channel dimension, the downsampled feature maps are sequentially upsampled to the same size using nearest neighbor interpolation.
	
	\item Finally, the four sub-feature maps of the same size are subjected to the Concat operation and $1\times 1$ convolution operation of the channel dimension, respectively, and the generated results are multiplied with the feature map ${{\hat{X}}_{1}}$ to obtain the feature map ${{\hat{X}}_{2}}$.
\end{itemize}

The second part is expressed mathematically in the following equation.

\begin{equation}
	\begin{aligned}
	& \left[ {{X}_{1}},{{X}_{2}},{{X}_{3}},{{X}_{4}} \right]=S\left( {{{\hat{X}}}_{1}} \right) \\ 
	& X_{1}^{c}=Conv3\times 3\left( {{X}_{1}} \right) \\ 
	& X_{i}^{c}=U\left( Conv3\times 3\left( M\left( {{X}_{i}} \right) \right) \right),1\le i\le 3 \\ 
	& {{X}^{c}}=Conv1\times 1\left( Concat\left( \left[ X_{1}^{c},X_{2}^{c},X_{3}^{c},X_{4}^{c} \right] \right) \right) \\ 
	& {{{\hat{X}}}_{2}}={{X}^{c}}*{{{\hat{X}}}_{1}}. 
	\end{aligned}
\end{equation}

Where $S$ denotes the channel division operation. $U$ represents the nearest neighbor interpolation upsampling operation. $M$ represents the maximum pooling of spatial dimensions.

One of the innovative designs in Part I and Part II is the use of feature map dot product, which abandons the traditional nonlinear activation function. The feature map dot product not only introduces nonlinear factors, but also improves the computational efficiency and feature interaction capability of the model.

\subsubsection{CFC Module}
It is well known that the effective extraction of local information plays a key role in further improving the quality of HR image reconstruction results. Therefore, this paper designs an improved CFC module based on the CCM module proposed in the literature \cite{ref206}. Compared with the CCM module, the CFC module designed in this paper discards the GELU activation function and uses the feature map dot product instead. The advantages of feature map dot product have been elaborated in the previous section.

The specific process of the CFC module is:

\begin{itemize}
	\item First, the input feature map $Y$ undergoes a normalization operation to generate the feature map $\hat{Y}$.
	
	\item The feature map $\hat{Y}$ is then subjected to a $3\times 3$-convolution operation to generate the feature map ${{\hat{Y}}_{0}}$. This operation is designed to encode spatially localized information, which implements the mixing of input feature channels and doubles the number of channels.
	
	\item Subsequently, the feature map ${{\hat{Y}}_{0}}$ undergoes channel bisection and feature map dot product operation to generate the feature map ${{\hat{Y}}_{1}}$. this operation not only improves the feature interaction capability of the model, but also further optimizes the computational efficiency of the model by reducing the number of channels.
	
	\item Finally, the feature map ${{\hat{Y}}_{1}}$ is subjected to a convolution operation $1\times 1$ to generate the feature map ${{\hat{Y}}_{2}}$. By this operation, the feature channels are further blended, while the number of channels is precisely adjusted to match the feature map $Y$.
\end{itemize}

The mathematical representation of the CFC module is given in the following equation.

\begin{equation}
	\begin{aligned}
		& {{{\hat{Y}}}_{0}}=Conv3\times 3\left( L\left( Y \right) \right) \\ 
		& \left[ {{Y}_{1}},{{Y}_{2}} \right]=S\left( {{{\hat{Y}}}_{0}} \right) \\ 
		& {{{\hat{Y}}}_{2}}=Conv1\times 1\left( {{Y}_{1}}*{{Y}_{2}} \right). 
	\end{aligned}
\end{equation}

\subsection{Global-Local Information Extraction Module}
Many details and textures in super-resolution images are localized, such as edges, corner points, texture patterns, etc. Deep learning models, especially Convolutional Neural Networks (CNNs), are able to accurately capture and recover these important local details by sliding a convolutional kernel over the image and learning a representation of the local region. The recovery of these details is crucial for improving the visual quality and fidelity of the image.

However, relying only on local information for super-resolution reconstruction may not be sufficient. Since the image contains rich global information, such as the overall structure, color distribution, and relationships among objects, in addition to local details. This information also plays a critical role in the process of accurate super-resolution image reconstruction. Therefore, in the super-resolution reconstruction process, in addition to utilizing local information, it is also necessary to combine global information to guide the reconstruction process. Global information helps the model to understand the content of the image more comprehensively, which in turn better guides the reconstruction of local details, thus generating higher quality super-resolution images.

Based on the above theory, this paper designs a global-local information extraction module that aims to enhance the reconstruction effect and quality of super-resolution images. By combining global and local information, this module aims to understand the image content more comprehensively and thus recover the details and textures in the image more accurately, with the ultimate goal of enhancing the visual quality of super-resolution images.

As shown in Fig. 3, the processing flow of the global-local information extraction module mainly contains the following steps:

\begin{figure*}[bht]
	\centering
	\includegraphics[width=6.5in]{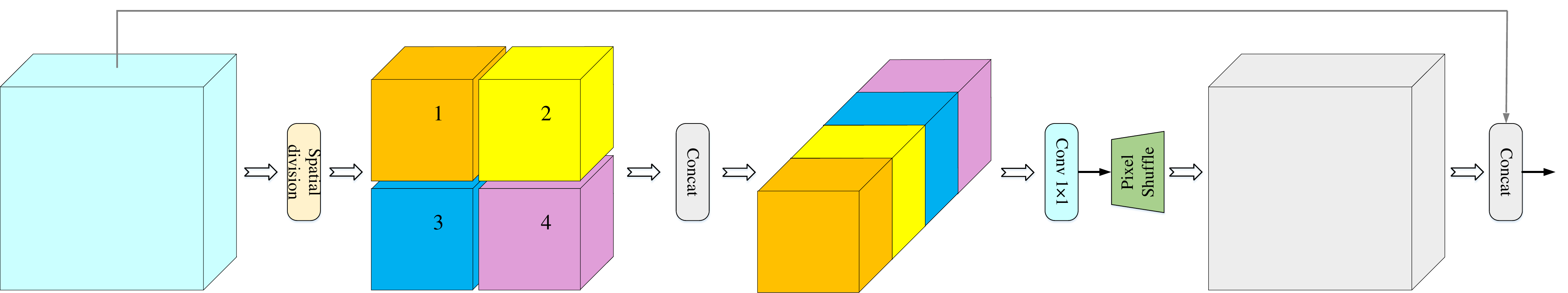}
	\caption{Diagram of global-local information extraction module.}
	\label{fig4}
\end{figure*}

\begin{itemize}
	\item First, the input size is $C\times H\times W$ feature map $Z$ is spatially divided to generate four sub-feature maps $\left[ {{Z}_{1}},{{Z}_{2}},{{Z}_{3}},{{Z}_{4}} \right]$. each sub-feature map maintains the same number of channels as the original feature map $Z$, but the spatial size is reduced to half of the original.
	
	\item The four sub-feature maps $\left[ {{Z}_{1}},{{Z}_{2}},{{Z}_{3}},{{Z}_{4}} \right]$ are then subjected to a channel-dimensional Concat operation to generate a $4C\times \left( {H}/{2}\; \right)\times \left( {W}/{2}\; \right)$ feature map of size $\hat{Z}$. This operation is designed to expand the receptive field in spatially localized regions, allowing the convolutional kernel to capture global contextual information without adding additional computational burden.
	
	\item Subsequently, the feature map $\hat{Z}$ undergoes $1\times 1$-convolution operation and PixelShuffle (PixelShuffle has better reconstruction effect and higher computational efficiency than other up-sampling methods) up-sampling operation to generate a feature map of size $C\times H\times W$ ${{\hat{Z}}_{1}}$. This operation not only skillfully realizes the fusion of the feature map channels, but also expands the spatial size of the feature maps, thus further enriching the The expression of image information is further enriched.
	
	\item Finally, the feature map ${{\hat{Z}}_{1}}$ and the feature map $Z$ undergo the Concat operation in channel dimension to produce the feature map ${{\hat{Z}}_{2}}$. This operation not only skillfully incorporates the residual information of the low-dimensional features, but also succeeds in effectively fusing the local information within the spatially small sensory field.
\end{itemize}

The mathematical formulation of the global-local information extraction module module is given in the following equation.

\begin{equation}
	\begin{aligned}
	& \left[ {{Z}_{1}},{{Z}_{2}},{{Z}_{3}},{{Z}_{4}} \right]=K\left( Z \right) \\ 
	& \hat{Z}=Concat\left( \left[ {{Z}_{1}},{{Z}_{2}},{{Z}_{3}},{{Z}_{4}} \right] \right) \\ 
	& {{{\hat{Z}}}_{2}}=Concat\left( \left[ PS\left( Conv1\times 1\left( {\hat{Z}} \right) \right),Z \right] \right).
	\end{aligned}
\end{equation}

where $K$ represents the spatial division. $PS$ stands for PixelShuffle operation.

\subsection{Loss Function}
In order to evaluate the model performance more comprehensively, this paper adopts the loss function from the literature [206], which incorporates the mean absolute error loss and the frequency loss based on the Fast Fourier Transform. This loss function is defined as follows:

\begin{equation}
	L=\left\| {{I}_{SR}}-{{I}_{HR}} \right\|+\gamma \left\| \varphi \left( {{I}_{SR}} \right)-\varphi \left( {{I}_{HR}} \right) \right\|.
\end{equation}

Where ${{I}_{SR}}$ represents the reconstructed super-resolution image, ${{I}_{HR}}$ is the corresponding super-resolution ground truth, and $\gamma =0.05$ is used as the weight coefficient. $\varphi $ is the fast Fourier transform (FFT).

Using the mean absolute error as the loss function ensures that the reconstructed image is closer to the original high-resolution image at the pixel level, thus improving the fidelity of the reconstructed image. Meanwhile, the frequency loss based on the Fast Fourier Transform helps the model to recover the detailed information in the image, which is crucial for improving the visual effect of the reconstructed image. Therefore, combining the loss function in the spatial and frequency domains can provide a more comprehensive evaluation metric, which helps the model to achieve better performance.

\section{Experiments}
In the experimental part, this paper conducts relevant comparative studies with a variety of current state-of-the-art classical super-resolution reconstruction algorithms, including EDSR \cite{ref138}, RCAN \cite{ref143}, HAN \cite{ref208}, SAFMN \cite{ref155} and SRFormer \cite{ref168}. Next, the experiments are described in detail, including qualitative and quantitative comparison tests performed on the five image datasets, as well as ablation analysis.

\subsection{Experimental Settings}
In the experiments, Adam optimizer is used to solve the proposed model where ${{\beta }_{1}}$ is set to 0.9 and ${{\beta }_{2}}$ is set to 0.99. The learning rate is gradually reduced from the initial $1\times {{10}^{-3}}$ to the minimum $1\times {{10}^{-5}}$ to optimize the training process. The entire model implementation and training process was done with the help of the PyTorch framework and was performed on NVIDIA GeForce RTX 3090 GPUs.

\subsection{Datasets}
The training dataset is derived from the DF2K dataset \cite{ref138}, which is a merger of DIV2K \cite{ref138} and Flickr2K \cite{ref211}.The DIV2K dataset consists of 800 training-set images, 100 validation-set images, and 100 test-set images, each of which has 2K resolution. In addition, the Flickr2K dataset has 2650 images, also in 2K size.

The test dataset consists of five benchmark datasets including Set5 \cite{ref212}, Set14 \cite{ref213}, B100 \cite{ref214}, Urban100 \cite{ref215} and Manga109 \cite{ref216}.

\subsection{Assessment Indicators}
In this paper, SSIM and PSNR are used as quantitative metrics for experimental results. And all the PSNR and SSIM values in this paper are calculated on the Y channel of the image converted to YCbCr color space.

\subsection{Performance Comparison}
In this section, in order to comprehensively evaluate the performance of the proposed model, this paper conducts exhaustive validation experiments on several benchmark datasets. Set5, Set14, B100, Urban100, and Manga109, in that order.These datasets contain a wide range of different types of images, which are able to fully reflect the performance of the model in different scenarios.

In addition, the validation experiments also explore the problem of super-resolution reconstruction at different scales in depth. Specifically, super-resolution reconstruction experiments are conducted at three different scales, namely $\times 2$, $\times 3$ and $\times 4$, in order to comprehensively evaluate the performance of the model at different magnifications.

\subsubsection{Quantitative Results}
Tables 1, 2 and 3 present the quantitative comparison results of different algorithms on five benchmark datasets and correspond to scales $\times 2$, $\times 3$ and $\times 4$ respectively. In addition, the complexity metrics of the different models are also presented in Tables 1, 2 and 3.

\begin{table*}[h]
	\caption{Quantitative results ($\times 2$ scales) of each algorithm on different test datasets, with larger values indicating better results. $\#$Params and $\#$FLOPs are metrics describing the complexity of the model and the size of the HR image is $1280\times 720$, with the red values representing the best results and the blue values the sub-optimal performance}
	\centering
	\begin{tabular}{c | c c c c c c}
		\hline
		Size&\multicolumn{6}{c}{$\times 2$}\\
		\hline
		Method & EDSR & RCAN & HAN & SAFMN & Ours & SRFormer \\
		$\#$Params [M] & 40.73 & 15.45 & 63.61 & \color{red}5.56 & \color{blue}5.57 & 10.38\\
		$\#$FLOPs [G] & 9387 & 3530 & 14551 & \color{blue}1274 & \color{red}1212 & 3179 \\
		Set5 & 38.11/0.9602& 38.27/0.9614 & 38.27/0.9614 & 38.28/0.9616 & \color{blue}{38.31/0.9617} & \color{red}{38.51/0.9627}\\
		Set14 & 33.92/0.9195 & 34.12/0.9216 & 34.16/0.9217 & 34.14/0.9220 & \color{blue}{34.17/0.9226} & \color{red}{34.44/0.9253}\\
		B100 & 32.32/0.9013 & 32.41/0.9027 & 32.41/\color{blue}{0.9027} & 32.39/0.9024 & \textcolor{blue}{32.41}/0.9026 & \color{red}{32.57/0.9046}\\
		Urban100 & 32.93/0.9351 & 33.34/0.9384 & \color{blue}{33.35/0.9385} & 33.06/0.9366 & 33.10/0.9369 & \color{red}{34.09/0.9449}\\
		Manga109 & 39.10/0.9773 & 39.44/0.9786 & 39.46/0.9785 & 39.56/0.9790 & \color{blue}{39.69/0.9792} & \color{red}{40.07/0.9802}\\
		\hline
	\end{tabular}
\end{table*}

\begin{table*}[h]
	\caption{Quantitative results ($\times 3$ scales) of each algorithm on different test datasets, with larger values indicating better results. $\#$Params and $\#$FLOPs are metrics describing the complexity of the model and the size of the HR image is $1280\times 720$, with the red values representing the best results and the blue values the sub-optimal performance}
	\centering
	\begin{tabular}{c | c c c c c c}
		\hline
		Size&\multicolumn{6}{c}{$\times 3$}\\
		\hline
		Method & EDSR & RCAN & HAN & SAFMN & Ours & SRFormer \\
		$\#$Params [M] & 43.68 & 15.63 & 64.35 & \color{red}5.58 & \color{blue}5.61 & 10.56\\
		$\#$FLOPs [G] & 4470 & 1586 & 6534 & \color{blue}569 & \color{red}548 & 1443 \\
		Set5 & 34.65/0.9280& 34.65/0.9280 & 34.75/0.9299 & \color{blue}34.80/0.9301 & {34.75/0.9290} & \color{red}{35.02/0.9323}\\
		Set14 & 30.52/0.8462 & 30.52/0.8462 & 30.67/0.8483 & 30.68/0.8485 & \color{blue}{30.69/0.8488} & \color{red}{30.94/0.8540}\\
		B100 & 29.25/0.8093 & 29.25/0.8093 & 29.32/{0.8110} & 29.34/0.8110 & \color{blue}{29.34}/0.8110 & \color{red}{29.48/0.8156}\\
		Urban100 & 28.80/0.8653 & 28.80/0.8653 & \color{blue}{29.10/0.8705} & 28.99/0.8679 & 28.98/0.8673 & \color{red}{30.04/0.8865}\\
		Manga109 & 34.17/0.9476 & 34.17/0.9476 & 34.48/0.9500 & 34.66/0.9504 & \color{blue}{34.75/0.9507} & \color{red}{35.26/0.9543}\\
		\hline
	\end{tabular}
\end{table*}

\begin{table*}[h]
	\caption{Quantitative results ($\times 4$ scales) of each algorithm on different test datasets, with larger values indicating better results. $\#$Params and $\#$FLOPs are metrics describing the complexity of the model and the size of the HR image is $1280\times 720$, with the red values representing the best results and the blue values the sub-optimal performance}
	\centering
	\begin{tabular}{c | c c c c c c}
		\hline
		Size&\multicolumn{6}{c}{$\times 4$}\\
		\hline
		Method & EDSR & RCAN & HAN & SAFMN & Ours & SRFormer \\
		$\#$Params [M] & 43.90 & 15.59 & 64.20 & \color{red}5.60 & \color{blue}5.66 & 10.52\\
		$\#$FLOPs [G] & 2895 & 918 & 3776 & \color{blue}321 & \color{red}311 & 842 \\
		Set5 & 32.46/0.8968& 32.63/0.9002 & 32.64/0.9002 & 32.65/0.9005 & \color{blue}{32.67/0.9005} & \color{red}{32.93/0.9041}\\
		Set14 & 28.80/0.7876 & 28.87/0.7889 & 28.90/0.7890 & 28.96/0.7898 & \color{blue}{28.96/0.7902} & \color{red}{29.08/0.7953}\\
		B100 & 27.71/0.7420 & 27.77/0.7436 & 27.80/{0.7442} & 27.82/0.7440 & \color{blue}{27.82}/0.7442 & \color{red}{27.94/0.7520}\\
		Urban100 & 26.64/0.8033 & 26.82/0.8087 & 26.85/\textcolor{blue}{0.8094} & 26.81/0.8058 & \textcolor{blue}{26.86}/0.8070 & \color{red}{27.68/0.8311}\\
		Manga109 & 31.02/0.9148 & 31.22/0.9173 & 31.42/0.9177 & 31.59/0.9192 & \color{blue}{31.64/0.9198} & \color{red}{32.21/0.9271}\\
		\hline
	\end{tabular}
\end{table*}

Due to the efficient structural design of the model in this paper, its complexity is comparable to that of the latest CNN-based algorithm, SAFMN, but it shows a significant advantage in terms of performance. Compared with other models, the model in this paper achieves a significant reduction in complexity. Specifically, as seen from the $\times 2$-scale results in Table 1, the model in this paper has 46$\%$ fewer parameters and 62$\%$ less computation compared to SRFormer; 91$\%$ fewer parameters and 92$\%$ less computation compared to HAN; 64$\%$ fewer parameters and 66$\%$ less computation compared to RCAN; and 86$\%$ fewer parameters and 86$\%$ less computation compared to EDSR. amount was reduced by 87$\%$. These data show that the model in this paper significantly reduces the computational complexity and the number of parameters while maintaining high performance.

The data presented in Tables 1, 2 and 3 show that the algorithm in this paper exhibits higher accuracy compared to the SAFMN algorithm with comparable complexity. Meanwhile, compared with other CNN and Transformer-based methods, this paper's algorithm has a significant advantage in model efficiency, and its reconstruction performance is also competitive. Therefore, considering both model complexity and accuracy, this paper's algorithm undoubtedly achieves the optimal results.

\subsubsection{Qualitative Results}
In addition to the quantitative assessment, qualitative comparisons are also performed in this paper. Fig. 4 illustrates the results of the visual comparison on the Urban100 dataset for the three scales $\times 2$, $\times 3$ and $\times 4$. By observing Fig. 4, it can be found that the visual results of the model in this paper are basically comparable to those of SRFormer and outperform the other models.

\begin{figure*}[h]
	\centering
	\includegraphics[width=5.5in]{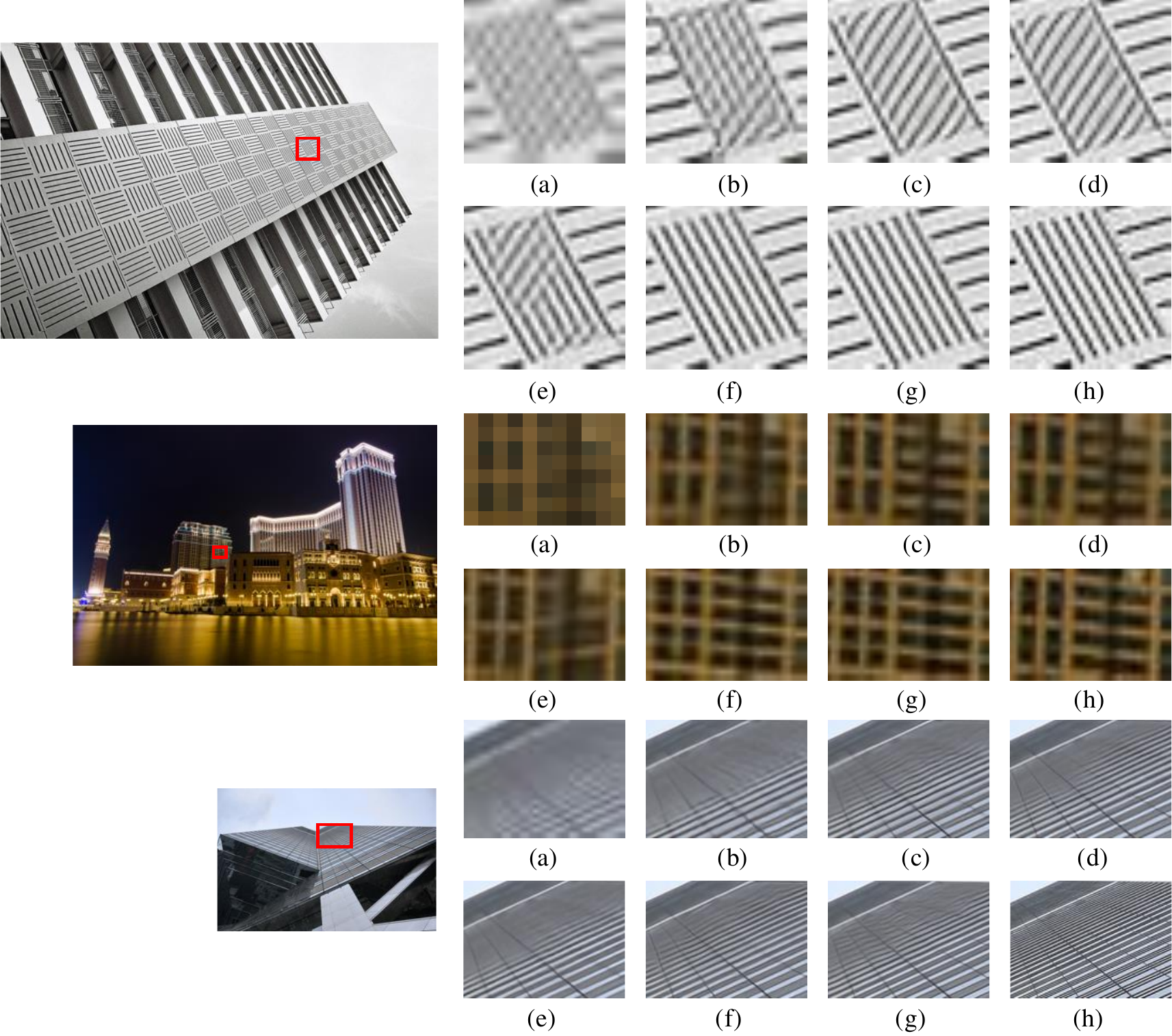}
	\caption{Qualitative comparison results of different algorithms on the Urban100 dataset, the first image is the $\times 2$-scale result, the second image is the $\times 3$-scale result, and the third image is the $\times 4$-scale result. (a) LR images, (b) EDSR results, (c) RCAN results, (d) HAN results, (e) SAFMN results, (f) results from this chapter, (g) SRFormer results, (h) HR images.}
	\label{fig6}
\end{figure*}

\subsection{Ablation Analysis}
In this paper, ablation experiments are implemented on Urban100 and Manga109 datasets to verify the performance of each module. Three modules are specifically included: the Spatial Channel Adaptive Modulation (SCAM) module, the Hybrid Channel Convolution (CFC) module, and the Global-Local Information Extraction module.

The ablation study is shown below:

\begin{itemize}
	\item -w/o CFC, the model removes only the CFC module;
	
	\item -w/o SCAM, the model removes only the SCAM module;
	
	\item -w/o Global-Local Information Extraction module, the model removes only the Global-Local Information Extraction module.
\end{itemize}

\subsubsection{Qualitative Results} 
Fig. 5 visualizes the qualitative results of different components on the Urban100 dataset and the Manga109 dataset at different scales. By observing Fig. 5, it can be clearly seen that the model containing all components performs optimally in terms of reconstruction results, presenting a clearer and more realistic visual effect.

\begin{figure*}[h]
	\centering
	\includegraphics[width=5in]{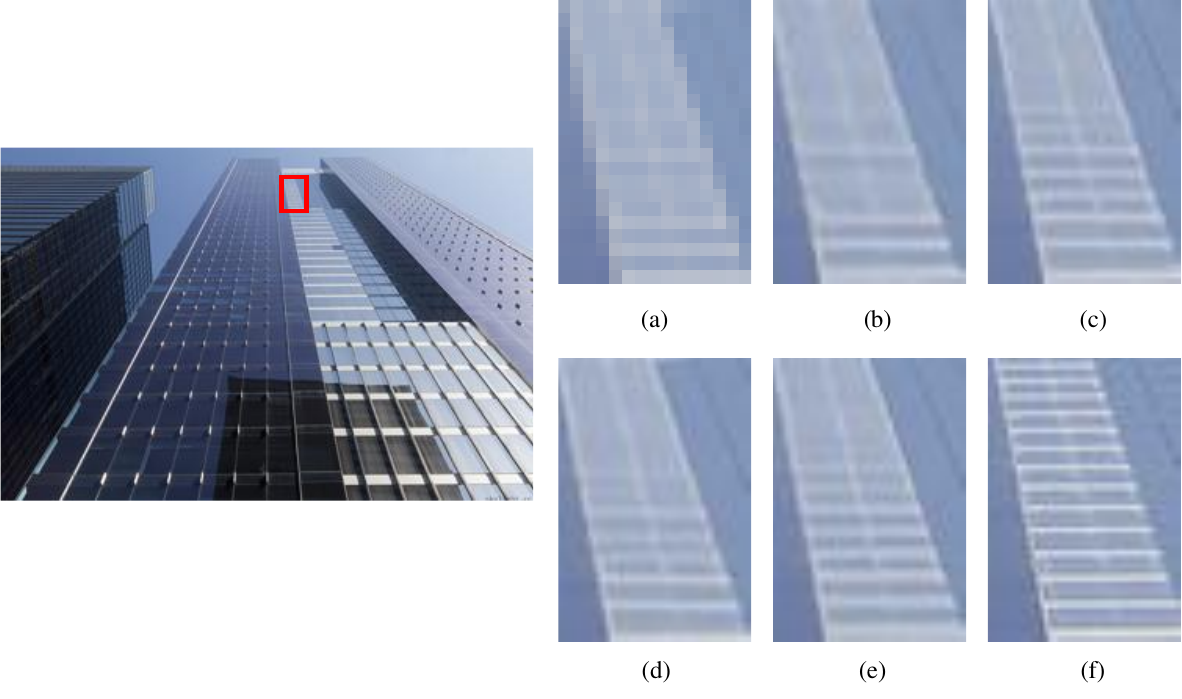}
	\caption{Qualitative comparison results of different components on Urban100 dataset ($\times 3$ scales). (a) LR image, (b) -w/o CFC results, (c) -w/o SCAM results, (d) -w/o Global-Local Information Extraction Module results, (e) results with all components included, (f) HR image}
	\label{fig9}
\end{figure*}

\subsubsection{Quantitative Results}

Table 4 details the quantitative results of the different components on the Urban100 dataset versus the Manga109 dataset at $\times 3$ scales. Since the CFC module and the SCAM module are the key components that make up the basic Block module and the basic Block module occupies a large portion of the model structure in this chapter, removing the CFC module and the SCAM module, respectively, will have a significant impact on the model performance. By carefully analyzing the data in Table 4, it can be found that the model with all components included achieves the optimal performance in all performance indicators.

\begin{table*}[!t]
	\setlength\tabcolsep{4pt} 
	\caption{Quantitative results for each component on the Urban100 dataset and the Manga109 dataset ($\times 3$-scale), with larger values indicating better results}
	\centering
	\label{tab:UGAN and UIEBD}
	\begin{tabular}{ccc}
		\hline
		Distillation component&Urban100 & Manga109 \\
		\hline
		-w/o CFC&28.33/0.8328&33.89/0.9287\\
		-w/o SCAM&28.75/0.8634&34.50/0.9493\\
		-w/o Global-Local Information Extraction Module&28.87/0.8652&34.59/0.9499\\
		all components included & 28.98/0.8673 & 34.75/0.9507\\
		\hline
	\end{tabular}
\end{table*}

\section{Conclusion}
In this paper, a generalized super-resolution reconstruction algorithm is unearthed. The core of the algorithm lies in the clever integration of the global-local information extraction module and the basic block module, which work together to realize the reconstruction of high-quality images.

The global-local information extraction module is able to capture all kinds of information in the image in a comprehensive and in-depth manner, whether it is global structural features or local texture details, all of which can be accurately extracted. This module provides rich and powerful information support for the subsequent reconstruction process, making the reconstruction results more accurate and delicate. The basic Block module is another core in the algorithm. It combines the two techniques of spatial channel adaptive modulation and hybrid channel convolution, which enhances the flexibility of the algorithm and improves the efficiency of feature extraction. This module makes the algorithm more flexible to adapt to all kinds of image characteristics, thus enhancing the reconstruction effect.

Considering the two key factors of reconstruction accuracy and model complexity, the algorithm in this paper achieves the optimal result in terms of comprehensive performance. This fully proves the advancement and practicability of the algorithm, and opens up new ideas and directions for the research in the field of super-resolution reconstruction.

\section*{Acknowledgments}
There is no financial support for this work. The authors declare that no conflicts of interest exist.



\begin{thebibliography}{1}
\bibliographystyle{IEEEtran}


\bibitem{ref115}
Gerchberg R W. Super-resolution through error energy reduction[J]. Optica Acta: International Journal of Optics, 1974, 21(9): 709-720.
\bibitem{ref116}
Kopf J, Cohen M F, Lischinski D, et al. Joint bilateral upsampling[J]. ACM Transactions on Graphics, 2007, 26(3): 96-es.
\bibitem{ref117}
He K, Sun J, Tang X. Guided image filtering[J]. IEEE transactions on pattern analysis and machine intelligence, 2012, 35(6): 1397-1409.
\bibitem{ref118}
Ham B, Cho M, Ponce J. Robust image filtering using joint static and dynamic guidance[C]. In: Proceedings of the IEEE Conference on Computer Vision and Pattern Recognition (CVPR). 2015: 4823-4831.
\bibitem{ref119}
Li Y, Min D, Do M N, et al. Fast guided global interpolation for depth and motion[C]. In: European Conference on Computer Vision (ECCV). 2016: 717-733.
\bibitem{ref120}
Yang S, Wang Z, Zhang L, et al. Dual-geometric neighbor embedding for image super resolution with sparse tensor[J]. IEEE transactions on image processing, 2014, 23(7): 2793-2803.
\bibitem{ref121}
Deng C, Xu J, Zhang K, et al. Similarity constraints-based structured output regression machine: An approach to image super-resolution[J]. IEEE transactions on neural networks and learning systems, 2015, 27(12): 2472-2485.
\bibitem{ref156}
Ledig C, Theis L, Huszár F, et al. Photo-realistic single image super-resolution using a generative adversarial network[C]. In: Proceedings of the IEEE/CVF Conference on Computer Vision and Pattern Recognition (CVPR). 2017: 105-114.
\bibitem{ref157}
Wang X, Yu K, Wu S, et al. ESRGAN: Enhanced super-resolution generative adversarial networks[C]. In: European Conference on Computer Vision Workshops (ECCVW). 2018: 63-79.
\bibitem{ref158}
Sajjadi M S M, Scholkopf B, Hirsch M. Enhancenet: Single image super-resolution through automated texture synthesis[C]. In: Proceedings of the IEEE International Conference on Computer Vision (ICCV). 2017: 4491-4500.
\bibitem{ref159}
Park S J, Son H, Cho S, et al. Srfeat: Single image super-resolution with feature discrimination[C]. In: European conference on computer vision (ECCV). 2018: 439-455.
\bibitem{ref160}
Pan X, Zhan X, Dai B, et al. Exploiting deep generative prior for versatile image restoration and manipulation[J]. IEEE Transactions on Pattern Analysis and Machine Intelligence, 2021, 44(11): 7474-7489.
\bibitem{ref161}
Wei Y, Gu S, Li Y, et al. Unsupervised real-world image super resolution via domain-distance aware training[C]. In: Proceedings of the IEEE/CVF Conference on Computer Vision and Pattern Recognition (CVPR). 2021: 13385-13394
\bibitem{ref162}
Ma C, Rao Y, Cheng Y, et al. Structure-preserving super resolution with gradient guidance[C]. In: Proceedings of the IEEE/CVF Conference on Computer Vision and Pattern Recognition (CVPR). 2020: 7769-7778.
\bibitem{ref163}
Liang J, Zeng H, Zhang L. Details or artifacts: A locally discriminative learning approach to realistic image super-resolution[C]. In: Proceedings of the IEEE/CVF Conference on Computer Vision and Pattern Recognition (CVPR). 2022: 5657-5666.
\bibitem{ref164}
Han D, Kim J, Kim J. Deep pyramidal residual networks[C]. In: Proceedings of the IEEE/CVF Conference on Computer Vision and Pattern Recognition (CVPR). 2017: 5927-5935.
\bibitem{ref165}
Liang J, Cao J, Sun G, et al. Swinir: Image restoration using swin transformer[C]. In: Proceedings of the IEEE/CVF International Conference on Computer Vision (ICCV). 2021: 1833-1844.
\bibitem{ref166}
Liu Z, Lin Y, Cao Y, et al. Swin transformer: Hierarchical vision transformer using shifted windows[C]. In: Proceedings of the IEEE/CVF International Conference on Computer Vision (ICCV). 2021: 10012-10022.
\bibitem{ref167}
Zhang X, Zeng H, Guo S, et al. Efficient long-range attention network for image super-resolution[C]. In: European Conference on Computer Vision. Cham: Springer Nature Switzerland, 2022: 649-667.
\bibitem{ref168}
Zhou Y, Li Z, Guo C, et al. SRFormer: Permuted self-attention for single image super-resolution[C]. In: Proceedings of the IEEE/CVF International Conference on Computer Vision (ICCV). 2023.
\bibitem{ref128}
Dong C, Loy C C, He K, et al. Learning a deep convolutional network for image super-resolution[C]. In: European Conference on Computer Vision (ECCV). 2014: 184-199.
\bibitem{ref129}
Dong C, Loy C C, He K, et al. Image super-resolution using deep convolutional networks[J]. IEEE Transactions on Pattern Analysis and Machine Intelligence, 2015, 38(2): 295-307.
\bibitem{ref130}
Simonyan K, Zisserman A. Very deep convolutional networks for large-scale image recognition[C]. In: 3rd International Conference on Learning Representations (ICLR). 2015.
\bibitem{ref131}
Kim J, Lee J K, Lee K M. Accurate image super-resolution using very deep convolutional networks[C]. In: Proceedings of the IEEE Conference on Computer Vision and Pattern Recognition (CVPR). 2016: 1646-1654.
\bibitem{ref132}
Mao X J, Shen C, Yang Y B. Image restoration using very deep convolutional encoder-decoder networks with symmetric skip connections[C]. In: Proceedings of the 30th International Conference on Neural Information Processing Systems (NeurIPS). 2016: 2810-2818.
\bibitem{ref133}
Kim J, Lee J K, Lee K M. Deeply-recursive convolutional network for image super-resolution[C]. In: Proceedings of the IEEE Conference on Computer Vision and Pattern Recognition (CVPR). 2016: 1637-1645.
\bibitem{ref134}
Tai Y, Yang J, Liu X. Image super-resolution via deep recursive residual network[C]. In: Proceedings of the IEEE Conference on Computer Vision and Pattern Recognition (CVPR). 2017: 3147-3155.
\bibitem{ref135}
Tai Y, Yang J, Liu X, et al. Memnet: A persistent memory network for image restoration[C]. In: Proceedings of the IEEE International Conference on Computer Vision (ICCV). 2017: 4539-4547.
\bibitem{ref136}
Dong C, Loy C C, Tang X. Accelerating the super-resolution convolutional neural network[C]. In: European Conference on Computer Vision (ECCV). 2016: 391-407.
\bibitem{ref137}
Shi W, Caballero J, Huszár F, et al. Real-time single image and video super-resolution using an efficient sub-pixel convolutional neural network[C]. In: Proceedings of the IEEE Conference on Computer Vision and Pattern Recognition (CVPR). 2016: 1874-1883.
\bibitem{ref138}
Lim B, Son S, Kim H, et al. Enhanced deep residual networks for single image super-resolution[C]. In: Proceedings of the IEEE Conference on Computer Vision and Pattern Recognition Workshops (CVPRW). 2017: 136-144.
\bibitem{ref139}
Ahn N, Kang B, Sohn K A. Fast, accurate, and lightweight super-resolution with cascading residual network[C]. In: European Conference on Computer Vision (ECCV). 2018: 252-268.
\bibitem{ref140}
Tong T, Li G, Liu X, et al. Image super-resolution using dense skip connections[C]. In: Proceedings of the IEEE International Conference on Computer Vision (ICCV). 2017: 4799-4807.
\bibitem{ref141}
Zhang Y, Tian Y, Kong Y, et al. Residual dense network for image super-resolution[C]. In: Proceedings of the IEEE Conference on Computer Vision and Pattern Recognition (CVPR). 2018: 2472-2481.
\bibitem{ref142}
Huang G, Liu Z, Van Der Maaten L, et al. Densely connected convolutional networks[C]. In: Proceedings of the IEEE Conference on Computer Vision and Pattern Recognition (CVPR). 2017: 4700-4708.
\bibitem{ref143}
Zhang Y, Li K, Li K, et al. Image super-resolution using very deep residual channel attention networks[C]. In: European Conference on Computer Vision (ECCV). 2018: 286-301.
\bibitem{ref144}
Mei Y, Fan Y, Zhou Y, et al. Image super-resolution with cross-scale non-local attention and exhaustive self-exemplars mining[C]. In: Proceedings of the IEEE/CVF Conference on Computer Vision and Pattern Recognition (CVPR). 2020: 5690-5699.
\bibitem{ref145}
Mei Y, Fan Y, Zhou Y. Image super-resolution with non-local sparse attention[C]. In: Proceedings of the IEEE/CVF Conference on Computer Vision and Pattern Recognition (CVPR). 2021: 3517-3526.
\bibitem{ref146}
Magid S A, Zhang Y, Wei D, et al. Dynamic high-pass filtering and multi-spectral attention for image super-resolution[C]. In: Proceedings of the IEEE/CVF Conference on Computer Vision and Pattern Recognition (CVPR). 2021: 4288-4297.
\bibitem{ref147}
Wang Z, Liu D, Yang J, et al. Deep networks for image super-resolution with sparse prior[C]. In: Proceedings of the IEEE International Conference on Computer Vision (ICCV). 2015: 370-378.
\bibitem{ref148}
Yang J, Wright J, Huang T S, et al. Image super-resolution via sparse representation[J]. IEEE transactions on image processing, 2010, 19(11): 2861-2873.
\bibitem{ref149}
Lai W S, Huang J B, Ahuja N, et al. Deep laplacian pyramid networks for fast and accurate super-resolution[C]. In: Proceedings of the IEEE/CVF Conference on Computer Vision and Pattern Recognition (CVPR). 2017: 624-632.
\bibitem{ref150}
Lai W S, Huang J B, Ahuja N, et al. Fast and accurate image super-resolution with deep laplacian pyramid networks[J]. IEEE Transactions on Pattern Analysis and Machine Intelligence, 2018, 41(11): 2599-2613.
\bibitem{ref151}
Haris M, Shakhnarovich G, Ukita N. Deep back-projection networks for super-resolution[C]. In: Proceedings of the IEEE/CVF Conference on Computer Vision and Pattern Recognition (CVPR). 2018: 1664-1673.
\bibitem{ref152}
Haris M, Shakhnarovich G, Ukita N. Recurrent back-projection network for video super-resolution[C]. In: Proceedings of the IEEE/CVF Conference on Computer Vision and Pattern Recognition (CVPR). 2019: 3897-3906.
\bibitem{ref153}
Li Z, Yang J, Liu Z, et al. Feedback network for image super-resolution[C]. In: Proceedings of the IEEE/CVF Conference on Computer Vision and Pattern Recognition (CVPR). 2019: 3867-3876.
\bibitem{ref154}
Kong F, Li M, Liu S, et al. Residual local feature network for efficient super-resolution[C]. In: Proceedings of the IEEE/CVF Conference on Computer Vision and Pattern Recognition Workshops (CVPRW). 2022: 766-776.
\bibitem{ref155}
Sun L, Dong J, Tang J, et al. Spatially-Adaptive Feature Modulation for Efficient Image Super-Resolution[C]. In: Proceedings of the IEEE International Conference on Computer Vision (ICCV). 2023.
\bibitem{ref206}
Li Y, Chen N, Zhang J. Fast and high sensitivity focusing evaluation function[J]. Application Research of Computers, 2010, 27(4): 1534-1536.
\bibitem{ref208}
Niu B, Wen W, Ren W, et al. Single image super-resolution via a holistic attention network[C]. In: European Conference on Computer Vision (ECCV). 2020: 191-207.
\bibitem{ref211}
Timofte R, Agustsson E, Van Gool L, et al. Ntire 2017 challenge on single image super-resolution: Methods and results[C]. In: Proceedings of the IEEE Conference on Computer Vision and Pattern Recognition Workshops (CVPRW). 2017: 114-125.
\bibitem{ref212}
Bevilacqua M, Roumy A, Guillemot C, et al. Low-complexity single-image super-resolution based on nonnegative neighbor embedding[C]. In: Proceedings of the 23rd British Machine Vision Conference (BMVC). 2012: 1-10.
\bibitem{ref213}
Zeyde R, Elad M, Protter M. On single image scale-up using sparse-representations[C]. In: Curves and Surfaces, 2012: 711-730.
\bibitem{ref214}
Arbelaez P, Maire M, Fowlkes C, et al. Contour detection and hierarchical image segmentation[J]. IEEE Transactions on Pattern Analysis and Machine Intelligence, 2010, 33(5): 898-916.
\bibitem{ref215}
Huang J B, Singh A, Ahuja N. Single image super-resolution from transformed self-exemplars[C]. In: Proceedings of the IEEE/CVF Conference on Computer Vision and Pattern Recognition (CVPR). 2015: 5197-5206.
\bibitem{ref216}
Matsui Y, Ito K, Aramaki Y, et al. Sketch-based manga retrieval using manga109 dataset[J]. Multimedia Tools and Applications, 2017, 76: 21811-21838.

\end{thebibliography}
%

\end{document}